\makeatletter
\let\NAT@parse\undefined
\makeatother

\newcommand{\sysname}{\textsc{Coopernaut}\xspace}
\newcommand{\envname}{\textsc{AutoCastSim}\xspace}

\newcommand{\ie}{\textit{i.e.,}\xspace}
\newcommand{\eg}{\textit{e.g.,}\xspace}

\documentclass[10pt,twocolumn,letterpaper]{article}
\usepackage{cvpr}              % To produce the CAMERA-READY version
\usepackage{graphicx}
\usepackage{amsmath}
\usepackage{amssymb}
\usepackage{booktabs}
\usepackage[pagebackref,breaklinks,colorlinks]{hyperref}
\usepackage{graphicx}
\usepackage{caption}
\usepackage[dvipsnames]{xcolor}
\usepackage{subcaption}
\usepackage{bbm}
\usepackage[numbers]{natbib}
\usepackage[ruled,vlined]{algorithm2e}
\usepackage{amsmath}
\usepackage{hyperref}
\usepackage{natbib}
\usepackage{multirow}
\usepackage{float}
\usepackage{cellspace, tabularx}
\usepackage[accsupp]{axessibility}  % Improves PDF readability for those with disabilities.

\usepackage[capitalize]{cleveref}
\crefname{section}{Sec.}{Secs.}
\Crefname{section}{Section}{Sections}
\Crefname{table}{Table}{Tables}
\crefname{table}{Tab.}{Tabs.}

\def\cvprPaperID{6482} % *** Enter the CVPR Paper ID here
\def\confName{CVPR}
\def\confYear{2022}

\title{\LARGE \bf
\sysname: End-to-End Driving with\\Cooperative Perception for Networked Vehicles
}

\author{
Jiaxun Cui$^{*1}$
\quad 
Hang Qiu$^{*2}$
\quad 
Dian Chen$^{1}$
\quad 
Peter Stone$^{1,3}$
\quad Yuke Zhu$^{1}$ 
\\
$^1$The University of Texas at Austin
\quad
$^2$Stanford University
\quad
$^3$Sony AI\\
{\small cuijiaxun@utexas.edu, hangqiu@stanford.edu, \{dchen, pstone, yukez\}@cs.utexas.edu}
}

\begin{document}
\maketitle
\def\thefootnote{$^*$}\footnotetext{Equal contribution. Correspondence: \url{cuijiaxun@utexas.edu}, \url{hangqiu@stanford.edu}, \url{yukez@cs.texas.edu}}
%===============================================================================

\begin{abstract}
    Optical sensors and learning algorithms for autonomous vehicles have dramatically advanced in the past few years. Nonetheless, the reliability of today's autonomous vehicles is hindered by the limited line-of-sight sensing capability and the brittleness of data-driven methods in handling extreme situations.
    % have improved in recent years, moving objects occluded to the egocentric sensors can still result in emergency situations, in which very little time is left for autonomous vehicles to react, thus causing traffic accidents. 
    With recent developments of telecommunication technologies, cooperative perception with vehicle-to-vehicle communications has become a promising paradigm to enhance autonomous driving in dangerous or emergency situations. We introduce \sysname, an end-to-end learning model that uses cross-vehicle perception for vision-based cooperative driving. Our model encodes LiDAR information into compact point-based representations that can be transmitted as messages between vehicles via realistic wireless channels.
    % and shares LiDAR information, and translates the 
    % \jiaxun {(Need explanation)joint representations} \changed{agglomerate representations of perception and translate them} to control signals. This model only requires realistic communication bandwidth, and is \jiaxun{less (than what)  sensitive to} \changed{agnostic to} the number of neighborhood agents because of the set operation \changed{operations on points?}. 
    % The intermediate representations learned are spatially-aware, physically meaningful for spatial sharing, and 
    % \jiaxun{are able to keep the resolution (need explanation)} \changed{are able to avoid redundant voxelization steps} compared to voxel models. 
    To evaluate our model, we develop \envname, a network-augmented driving simulation framework with example accident-prone scenarios. Our experiments on \envname suggest that our cooperative perception driving models lead to a 40\% improvement in average success rate over egocentric driving models in these challenging driving situations and a 5$\times$ smaller bandwidth requirement than prior work V2VNet.
    \sysname and \envname are available  at \url{https://ut-austin-rpl.github.io/Coopernaut/}.
    % We have made \sysname and \envname  publicly available at \url{https://ut-austin-rpl.github.io/Coopernaut/}.

    % great potential for an agent to learn a \jiaxun{safer and smarter than what} policy through vehicle-to-vehicle communication. 
    % \jiaxun{last sentence too generic}
    % \jiaxun{Talk about numbers improvement from eval results}
    
    % \changed{Our experiments suggest that the proposed vehicle-to-vehicle cooperative framework can produce safer driving policies than those only depends on egocentric sensors across studied scenarios.}
    
\end{abstract}

% \keywords{Autonomous Driving, Networked Vehicles, Cooperative Perception} 

%===============================================================================

%\section{Introduction}
	
%    Submission to CoRL 2021 will be entirely electronic, via a web site (not email). Information about the submission process and \LaTeX{} templates are available on the conference web site at \url{http://www.robot-learning.org/}. For camera ready submission, use the \texttt{final} option for the \texttt{\textbackslash usepackage} command. 

%===============================================================================

%\section{Citations}
%\label{sec:citations}

%	Citations can be made using either \textbackslash citep\{\} or \textbackslash citet\{\}, depending from the appropriateness. To avoid the citation moving to the next line, it is often a good practice to replace the space before with a tilde (\~{}) character.
%	Example 1: ``CoRL is the best conference ever~\citep{chen2020learning}.''
%	Example 2: ``\citet{chen2020learning} proved, both theoretically and numerically, th2d at CoRL is the best conference ever.''
	
%===============================================================================

% The maximum paper length is 8 pages excluding references and acknowledgements, and 10 pages including references and acknowledgements

\section{Introduction}
The widespread deployment of autonomous driving and advanced driver assistance systems is challenged by safety concerns. While deep learning has improved autonomy stacks with data-driven techniques~\cite{xu2017end,chen2015deepdriving,bojarski2016end}, learning-based driving policies to date are still brittle, especially in the face of extreme situations and corner cases that one might encounter only a few times every million miles of driving~\cite{blanco2016automated}. The lack of robustness of learning algorithms is exacerbated by the limited sensing capabilities of optical sensors on individual vehicles, such as stereo cameras and LiDAR, that are confined to line-of-sight sensing and unreliable in bad weather conditions. With the advent of new telecommunication technologies, such as 5G networks and vehicle-to-vehicle (V2V) communications, \emph{cooperative perception}~\cite{qiu2018avr, aoki2020cooperative,chen2019cooper,kim2015impact} is becoming a promising paradigm that enables sensor information to be shared between vehicles (and roadside devices~\cite{emp}) in real-time. The shared information can augment the field of view of individual vehicles and convey the intents and path plans of nearby vehicles, offering the potential to improve driving safety, particularly in accident-prone scenarios.
% \jiaxun{What is uncommon}

%Decisions from the ego sensors are far from enough, and many of the perceptive sensors are optical-based which can be occluded. For example, you are driving straight forward at an intersection on the green light, but there is a car running a red light. What makes this situation worse is that your left-side view is occluded by a left-turn vehicle in parallel with you, who is waiting for a chance to turn left. Whether there is a traffic violator is unclear to you. In this case, if you continue going forward based on your assumption that everyone observes the traffic rule, then you may be run over by the collider, or if you are too conservative, you will stop completely at a green light. So it is of vital importance to build vehicle-to-vehicle communication networks for multi-car observation sharing, to make the information for driving decisions more complete.

%Ideally, all the information can be shared via telecommunication technology, and your information processor is capable of dealing with so much data. But the truth is, even with the 5G communication, you still have a bottleneck of amount of information that can be transmitted in real-time. Our idea is that we learn an intermediate light-weight representation of the observations for each vehicle, and only transmit the representation to fit the current communication and computation capacity.

\begin{figure}
\centering
\includegraphics[width=\columnwidth]{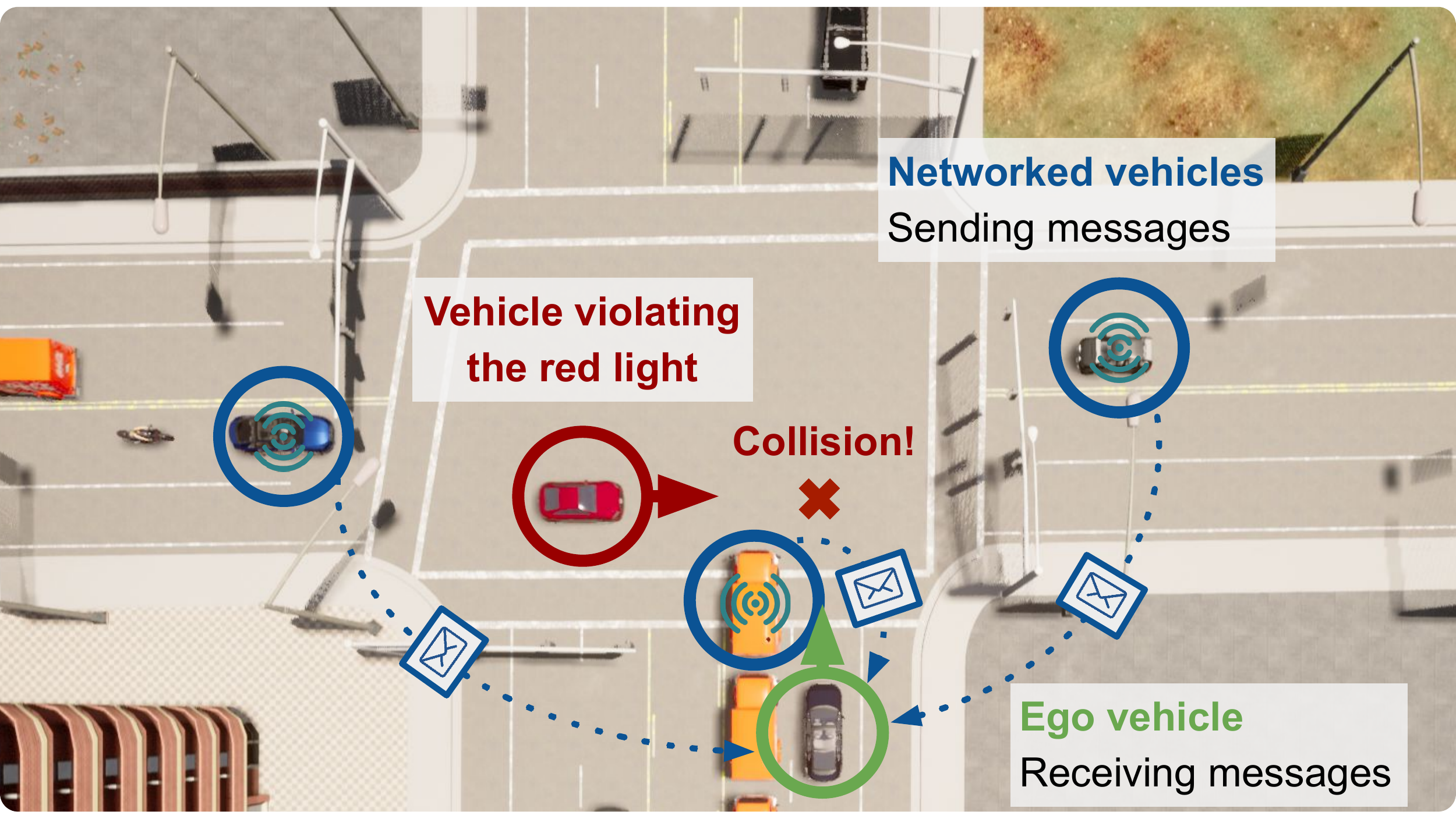}
% \vspace{-5mm}
\caption{\sysname enables vehicles to communicate critical information beyond occlusion and sensing range for vision-based driving. The blue dashed arrows are information sharing flows. Through cooperative perception, \sysname makes more informed driving decisions when line-of-sight sensing is limited.}
\vspace{-5mm}
\label{fig:bev_overview}
\end{figure} 

% \jiaxun{This section sounds like related work}
Ideally, learning autonomous driving policies with cooperative perception should take advantage of existing deep learning methods customized for ego perception~\cite{dosovitskiy2017carla,chen2020learning, prakash2021multi} by considering the combined sensory data from all vehicles as an \emph{augmented} version of on-board sensing. In practice, the efficacy of cooperative perception hinges on \emph{what} data to transmit within the limited network bandwidth and \emph{how} to use the aggregated information to build a coherent and accurate understanding of traffic situations.
%Prior work from the networking community has devised intelligent transmission strategies~\cite{qiu2018avr} to adaptively communicate sensor data between vehicles taking into account the channel variability. 
%\jiaxun{Did we really answer this question}
% Nonetheless, it remains an open question how to effectively fuse the sensory data from neighboring vehicles to improve driving performance. 
Recent work on cooperative driving has demonstrated the benefit of cross-vehicle perception for augmenting sensing capabilities and driving decisions~\cite{aoki2020cooperative,kim2015impact}.
%\yuke{[summarize existing work on cooperative perception]}. 
Nonetheless, these methods have abstracted away raw sensory data with low-dimensional meta-data.
% and addressed communication bandwidth limit. %\yuke{[describe what they have done]}. 
Prior work introduced 3D sensor fusion (AVR~\cite{qiu2018avr}, Cooper~\cite{chen2019cooper}) and representation fusion (V2VNet~\cite{wang2020v2vnet}) algorithms that aggregate perception results from nearby vehicles via V2V channels. They focused on 3D detection and motion forecasting on static datasets, rather than interactive driving policies.
% \jiaxun{We have definitely more differences from theirs}

% \textbf{Prior Work}
%There are also some works studying the perception aspect of networked vehicles [AVR, V2V, CMU]. They are motivated by the limited bandwidth of V-to-V communication and aim at a wiser choice of what to be transmitted. Nevertheless, they have not touched on how much improvement this communication network can bring to the control.
%On the other hand, though end-to-end driving policy learning is not a new topic in the autonomous driving field, with [..] and [..] establishing the work paradigms, they all only use ego sensor data.

%Many efforts in the multi-agent community have been devoted to making decisions on a joint observation from the neighbors. However, most of the work was conducted in the grid world or simple cases. We aim at solving scenarios which are more close to the real-world and deal with the 3D point cloud sensor data which contains more depth information. There are multiple representations of 3D data like voxel[..], sparse voxel[Minkowski] or point[PointNet++]. To fully utilize the sparse LiDAR structure, we are in favor of point based methods. PointTransformer[point transformer] is a natural choice for us to learn to encode the point clouds into several key points and then aggregate representations from a varied number of neighboring vehicles as transformer[transformer] is a set operation.

We introduce \sysname, an end-to-end cooperative driving model for networked vehicles. \sysname learns to fuse encoded LiDAR information shared by nearby vehicles under realistic V2V channel capacity. To communicate meaningful scene information from nearby vehicles while conforming to bandwidth limits, we design our driving policy architecture based on the Point Transformer~\cite{zhao2020point}, a self-attention network for point cloud processing.
% which has been shown to be remarkably effective at point cloud processing. %\yuke{[one phrase to talk about the advantages of PT over other methods]}.
This architecture pre-processes the raw point cloud, on each networked vehicle locally, into spatial-aware neural representations. These representations are compact, which can be efficiently transmitted over realistic wireless channels. Meanwhile, they are physically grounded, thus can be spatially transformed and aggregated with ego representations. The entire architecture is end-to-end differentiable, permitting control supervision (imitating an oracle planner with access to privileged information) to flow back to the perception stack, thus ensuring the learned representations and messages contain task-relevant information.
%
%Nonetheless, learning abstractions that are useful for control from the image or LiDAR is a nontrivial task, while most of the prior work only has limited supervision data. [LbC] provides a nice solution by first developing an automated expert policy who has access to the ground truth data in the simulator. We follow the learning structure of LbC to first design a perfect expert policy so that we can obtain endless expertise data when interacting with the environment. Then learn an homogeneous point cloud encoder, a cross-vehicle aggregation module of communicated data, and a control module based on the aggregated perceptual representations through the interactive data.

To examine the effectiveness of \sysname, we develop a CARLA-based simulation framework, \envname, where we designed three accident-prone scenarios. All the scenarios are designed to be challenging for ego perception to fully comprehend the traffic situation. 
%\jiaxun{Long sentence, spatially meaningful explanation, awkward}
%\yuke{[Add 2-3 sentences to describe AutoCast and its main strengths, and the scenarios that we evaluate on]} 
%AutoCast has a built-in communication mechanism that allows customization and 4 scenarios to benchmark different control policies. We also provide a pipeline for data collection, imitation learning, and evaluation along with in AutoCast.
\envname has a built-in networking simulation for customizable multi-vehicle communications and an expert driving model with privileged information. We evaluate \sysname with voxel-based baselines~\cite{wang2020v2vnet} and different sensor fusion schemes.
% Our results 
%\yuke{[Once we have them, summarize the key findings and mention the most impressive quantitative results]}
%Our experiments on AutoCast suggest that our cooperative perception driving models lead to an average of \changed{41.7\%} improvement over egocentric driving models in these challenging driving situations.
% We also provide a pipeline for data collection, imitation learning, and evaluation along with in AutoCast.

% \textbf{Problem Statement}
In summary, our main contributions are as follows: %we investigate end-to-end policy learning for networked autonomous vehicles. To our knowledge, this is the first work on end-to-end cooperative perception for vision-based driving policy learning. 
\begin{itemize}
\vspace{-1mm}
\item We introduce \sysname, an end-to-end driving model with cooperative perception via V2V channels. Our model learns compact representations for communication that can be easily harnessed by the ego vehicle to improve its driving decisions.
\vspace{-2mm}
\item We develop a network-augmented autonomous driving simulation framework \envname to evaluate \sysname and baselines in accident-prone scenarios and to promote future research on vision-based cooperative perception.
\vspace{-2mm}
\item Our results show that \sysname substantially reduces  safety hazards for line-of-sight sensing. Its design improves both driving performance and communication efficiency over baselines.
\vspace{-1mm}
%    \item To our knowledge, we propose the first end-to-end autonomous driving control model using 3D cooperative perception
    % \item We developed an open-source network-augmented driving simulation framework, AutoCast, to evaluate cooperative perception in accident-prone scenarios;
    % \item  We introduced a point-based cooperative perception model, Cooperative Point Transformer, for extracting spatially meaningful and spatially transformable LiDAR information and aggregating multi-vehicle representations. \yuke{[a bit unclear what is ``spatial awareness'', ``transformer mechanism'', ``sensitivity''. think about better ways to summarize the key strengths of CPT] keeps the spatial awareness and leverages the transformer mechanism to reduce the sensitivity to the number of neighboring agents}.
    % \item Our model is end-to-end differentiable and is trained using imitation learning on algorithmic oracles. Our results demonstrates state-of-the-art danger avoidance capability with reasonable communication bandwidth compared to different schemes. \yuke{[be more concrete how our method is better] strongly justify the meaningfulness of cooperative perception}.

% with point-based multi-vehicle cooperative perception. There is an ego vehicle, along with its neighboring vehicles who are broadcasting their observations through the LiDAR by some communication protocols. 
 %
% The ego vehicle needs to aggregate the exchanged collective perception and its ego observation to make control decisions according to the joint information.
%
% \textbf{Contribution}
\end{itemize}
%
% Our CPT algorithm implementation and the AutoCast simulation framework will be released to the research community upon the publication of this work.

% \noindent
% We have made \sysname and \envname  publicly available at \url{https://ut-austin-rpl.github.io/Coopernaut/}.
% \vspace{-1mm}
%- we learn a spacial-aware message/representation for vehicle-to-vehicle communication that is physically meaningful, fit for spatial transformation, and informative for driving decision, 

%- we design a representation reasoner, fusing information from multiple vehicles to make better more informative driving decision.

%- open source simulator, benchmark, dataset

%- evaluate over diff schemes, prior work, state-of-the-art over performance baselines
\section{Related Work}
\vspace{-1mm}
%\begin{figure}[htbp]
%    \centering
%    \includegraphics[width=0.8\textwidth]{Figures/V2VNet.png}
%    \caption{V2VNet Overview}
%    \label{fig:v2v}
%\end{figure}
% \paragraph{Imitation Learning} has been the most successful approach for learning a driving policy. It was first developed in ALVINN~\cite{}. 
\paragraph{Deep Learning for Driving Policy.} Learning a driving controller involves training closed-loop policies using deep networks, usually via imitation learning and/or reinforcement learning. Imitation learning for autonomous driving was pioneered by~\citet{Pomerleau1988ALVINNAA}, and has since then been extended to urban and more complex scenarios~\cite{codevilla2018end,bansal2018chauffeurnet,sauer2018conditional,codevilla2019exploring,chen2020learning,Prakash2021CVPR}. Very recently, reinforcement learning has also made progress in autonomous driving~\cite{Toromanoff_2020_CVPR,chen2021learning}, showing potential to train better policies in complex situations~\cite{Toromanoff_2020_CVPR,chen2021learning}. However, reinforcement learning is known to be more data-hungry and requires engineering a high-quality reward function. We follow the imitation learning paradigm but use an expert oracle with complete global information~\cite{chen2020learning} for training efficiency.

\vspace{1mm}
\noindent\textbf{3D Perception for Autonomous Driving.} 
3D perception has become more popular in autonomous driving
% analyzes objects and geometries from 3D point cloud data. It has seen increasing popularity in autonomous driving, 
due to the decreasing cost of commoditized LiDAR sensors. \citet{Zhou2018VoxelNetEL} pioneered using 3D object detection in autonomous driving, and since then, it has been further developed as better models, and more advanced techniques have been discovered. Very recently, \citet{Prakash2021CVPR} also explored end-to-end driving using point cloud data. Two families of 3D perception backbones have been widely adopted: \textit{voxel}-based methods discretize points to voxels~\cite{Zhou2018VoxelNetEL,lang2019pointpillars,shi2020pv}; and \textit{point}-based methods directly operate on coordinates~\cite{qi2017pointnet,qi2017pointnet++,zhao2020point}. \sysname uses a transformer-based architecture \cite{vaswani2017attention} with point-based representations~\cite{zhao2020point,qi2017pointnet,qi2017pointnet++}, 
which preserves high spatial resolutions with discretization and requires lower bandwidths to transmit without compression needed by prior work~\cite{wang2020v2vnet}.

\vspace{1mm}
\noindent\textbf{Networked Vehicles and Cooperative Perception.}
Network connectivity offers a great potential for improving the safety and reliability of self-driving cars. Vehicles can now share surrounding information via Vehicle-to-vehicle (V2V) and vehicle-to-infrastructure (V2X) channels using wireless technologies, such as Dedicated Short Range Communication (DSRC)~\cite{DSRC} and cellular-assisted V2X (C-V2X)~\cite{LTE-Direct, Qualcomm:LTE-Direct}.
% , provide capabilities to exchange information among self-driving cars.
% by different types of transmission, \ie, multicasting, broadcasting, and unicasting. 
These V2V/V2X communication devices are increasingly deployed in current and upcoming vehicle models~\cite{Mercedes, Toyota}). The academic community has built city-scale wireless research platforms (COSMOS~\cite{cosmos}) and large connected vehicle testbeds (\eg, MCity~\cite{USTest}, DRIVE C2X~\cite{EuropeTest}), to explore the feasibility of cooperative vehicles and applications.
Prior work \cite{qiu2018avr, chen2019cooper} proposed cooperative perception systems that broaden the vehicle’s visual horizon by sharing raw visual information with other nearby vehicles. % 
Such systems can be scaled up to dense traffic scenarios leveraging edge servers~\cite{emp} or in an ad-hoc fashion~\cite{autocast}.
Recent work~\cite{wang2020v2vnet, disconet, xu2022opencood} proposed multi-agent perception models to process sensor information and share compact representations within a local traffic network.
In contrast, we focus on cooperative driving of networked vehicles with onboard visual data and realistic networking conditions, advancing towards real-world V2V settings.
\section{\sysname}
% This section first describes the problem setting of cooperative driving, then moves on to our proposed method and essential implementation details.

\subsection{Problem Statement}
Our goal is to learn a closed-loop policy that controls an autonomous ego vehicle, which receives LiDAR observations $O_{t}^{(\text{ego})}$ at time $t$. Assume that there exist a variable number of $N_t$ neighboring vehicles in the range of V2V communications at time $t$, where $O^{(i)}_t$ is the raw 3D point cloud from the onboard LiDAR of the $i$-th vehicle. The cooperative driving policy for the ego vehicle is to find a policy $\pi(a_t|O_{t}^{(\text{ego})},O^{(1)}_t,\ldots,O^{(N_t)}_t)$ that makes control decisions $a_t$ based on the joint observations of the ego vehicle and the $N_t$ neighboring vehicles. Here $\pi$ is parameterized by a deep neural network and trained end-to-end. In principle, we can transmit all cross-vehicle observations to the ego vehicle and process them as a whole. In practice, we have to take into account the networking bandwidth limit, which only allows for the message size orders of magnitude smaller. We thus first process the raw point clouds into compact representations, which can be transmitted through the V2V channels in real-time.

\subsection{Background: Point Transformer}
%\begin{figure}[h!]
%    \centering
%    \begin{subfigure}[b]{0.19\textwidth}
%        \centering
%        \includegraphics[width=\textwidth]{Figures/point_transformer_block.png}
%        \caption{point transformer block}
%        \label{fig:point_transformer_block}
%    \end{subfigure}
%    \begin{subfigure}[b]{0.28\textwidth}
%        \centering
%        \includegraphics[width=\textwidth]{Figures/transit%ion_down_block.png} 
%        \caption{transition down block}
%        \label{fig:transition_down_block}
%    \end{subfigure}
%    \caption{Point Transformer Modules}
%    \label{fig:point_transformer_layer}
%\end{figure}
Our model's backbone is the Point Transformer~\cite{zhao2020point}, a newly-developed neural network structure that learns compact point-based representations from 3D point clouds. It reasons about non-local interactions among points and produces permutation-invariant representations, making itself effective in aggregating multi-vehicle point clouds. Here we provide a brief review of Point Transformers.

We adopt the same design as \citet{zhao2020point}, which uses \emph{vector self-attention} to construct the Point Transformer Layer. We also apply subtraction between features and append a position encoding function $\delta$ to both the attention vector $\gamma$ and the transformed features $\alpha$:
\begin{equation}
    y_i=\sum_{x_j\in \mathcal{X}(i)}\rho(\gamma(\phi(x_i)-\psi(x_j)+\delta))\odot (\alpha(x_j)+\delta)
\end{equation}
Here the $x_i$ and $x_j$ are input features of the point $i$ and $j$ respectively, $y_i$ is the output attention feature for point $i$, and $\mathcal{X}(i)$ represents the set of points in the neighborhood of $x_i$; $\phi, \psi$ and $\alpha$ are point-wise feature transformations, an MLP; $\gamma$ is an MLP mapping function with two layers and one ReLU non-linearity; $\delta$ is a position-encoding function and $\rho$ is a normalization function \emph{softmax}. 
Given the 3D coordinates $p_i$, $p_j\in\mathbb{R}^3$ for point $i$ and $j$, the position-encoding function is formulated as follows:
\begin{equation}
    \delta=\theta(p_i-p_j)
\end{equation}
where $\theta$ is an MLP with two linear layers and one ReLU.
% non-linearity.

A \emph{Point Transformer block} is shown in Figure \ref{fig:method}, which integrates the self-attention layer, linear projections, and a residual connection. The input is a set of 3D points $p$ with a feature $x$ of each point. This block enables local information exchange among points, and produces new feature vectors for each point.
The \emph{down-sampling block} in Figure \ref{fig:method} is to reduce the cardinality of the point sets. We perform farthest point sampling \cite{eldar1997farthest} to the input set to obtain a well-spread subset, and then use kNN graph and (local) max pooling in the neighborhood to further condense the information to smaller sets of points. The output is a subset of the original input points with new features.

\subsection{Our Model}
%\begin{figure*}[!ht]
%    \centering
%    \includegraphics[width=0.9\textwidth]{Figures/cooperat%ive_point_transformer.png}
%    \caption{\sysname}
%    \label{fig:cpt}
%\end{figure*}
\begin{figure*}
    \centering
    \includegraphics[width=0.95\textwidth]{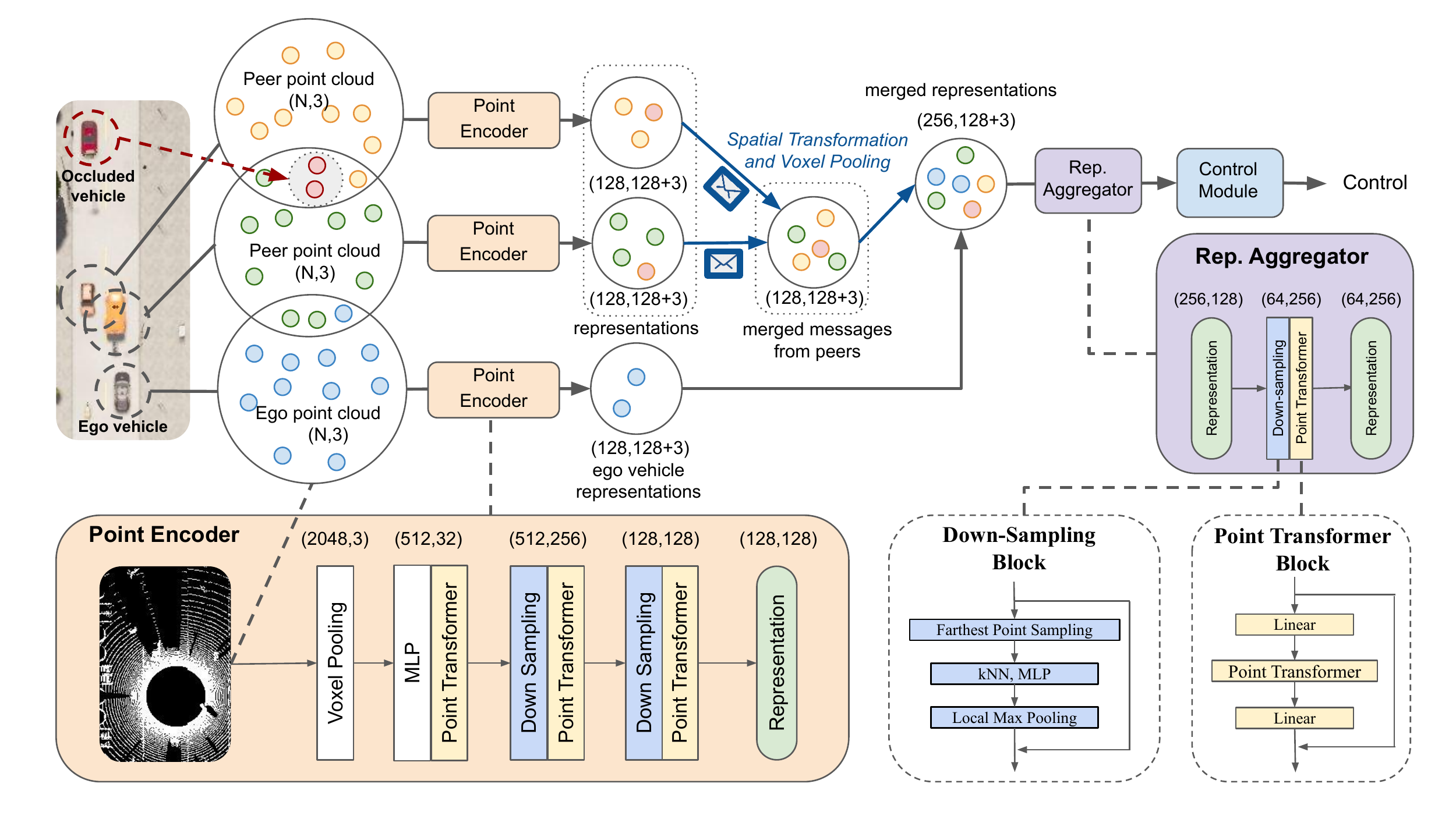}
    \vspace{-5mm}
    \caption{\sysname is an end-to-end vision-based driving model for networked vehicles. It contains a \textit{Point Encoder} to extract critical information locally for sharing, a \textit{Representation Aggregator} for merging multi-vehicle messages, and a \textit{Control Module} to reason about the joint messages. Each message produced by the encoder has 128 keypoint coordinates and their associated features. The message is then spatially transformed into the ego frame. The ego vehicle merges incoming messages and compute aggregated representations through voxel max-pooling. Finally, the aggregator synthesizes joint representations from the ego vehicle and all its neighbors before passing them to the Control Module to produce control decisions. The numbers in parentheses denote data dimensions.}
    %It spatially transform and aggregates the messages (point coordinates and representations) generated by the encoders, and make control decisions based on the merged messages. }
    \vspace{-4mm}
    \label{fig:method}
\end{figure*}

We use cross-vehicle perception to augment the sensing capabilities of the ego vehicle for it to make more informed decisions under challenging situations. The key challenge is to transmit sensory information efficiently through realistic V2V channels, understand the traffic situation from the aggregated information, and determine the reactive driving action in real-time.

% The key motivation for our problem is that occluded areas in the line-of-sight LiDAR scanning can be visible by another vehicle, but transmitting raw point cloud data requires an unrealistic bandwidth even with 5G. 
%So we need an \textit{point encoder} to generate compact representations of the critical information in the point cloud observation, and an \textit{representation aggregator} to incorporate every received message. To learn a driving policy, we also need a \textit{control module} to reason the aggregated messages. 

% \changed{A related V2V 3D detection and prediction work by \citet{wang2020v2vnet}} requires a 25Mbps bandwidth and needs voxelization steps, while we aim to achieve lower bandwidth requirements by operating directly on 3D points.

Our \sysname model, illustrated in Figure \ref{fig:method}, is composed of a Point Encoder for each neighboring V2V vehicle to encode its sensory data into compact messages, a Representation Aggregator to integrate the messages from neighboring cars with the ego perception, and a Control Module which translates the integrated representations to driving commands. %The entire architecture is fully differentiable and can be trained end-to-end. 
%Previous work \cite{wang2020v2vnet} requires a bandwidth of 25Mbps and needs voxelization steps, while we aim at achieve lower bandwidth requirements by operating directly on 3D points. The proposed model, \sysname (Figure \ref{fig:method}), is composed of a point cloud encoder for every individual vehicle to process LiDAR data, a cooperative perception module to aggregate the representations of neighboring cars and ego vehicle, and a control module which translates the agglomerated representations to control signals. This model is fully differentiable, so it can be trained in an end-to-end manner. 

\vspace{1mm}
\noindent\textbf{Point Encoder.}
%\begin{figure*}[h!]
%    \centering
%    \includegraphics[width=0.618\linewidth]{Figures/encode%r.png}
%    \caption{Encoder}
%    \label{fig:encoder}
%\end{figure*}
To reduce communication burdens, every V2V vehicle processes its own LiDAR data locally and encodes the raw 3D point clouds into keypoints, each associated with a compact representation learned by the Point Transformer blocks. We construct the encoder with three Point Transformer blocks accompanied by two down-sampling blocks, both with a downsampling rate of $(1,4,4)$. The final cardinality of intermediate representations is $P/16$, where $P$ is the number of points in the raw point cloud. In our experiments, we preprocess 65,536 raw LiDAR points to 2,048 points via voxel pooling, \ie, representing the points in a voxel grid using their voxel centroid.

The message $M_j$ produced by the $j$-th vehicle comprises a set of position-based representations $M_j$ and is mathematically described as $M_j={\{(p_{jk},R_{p_{jk}})\}}^K$,
% \begin{equation*}
%     \begin{split}
%         M_j&={\{R_{p_{jk}},p_{jk}\}}^K
%     \end{split}
% \end{equation*}
where $p_{jk}\in\mathbb{R}^3$ for $k=1,\ldots,K$ is the position of a keypoint in 3D space and $R_{p_{jk}}$ is its corresponding feature vector produced by the Point Encoder. We limit the size of $M_j$ to be at most $K$ tuples. These keypoints carrying features are in each vehicle's local frame. They preserve the spatial information as their coordinates are sampled from raw point clouds.
%they are of semantic meaning that they are key points with important information to be transferred for a more complete state representation in the ego vehicle's decision process.

\vspace{1mm}
\noindent\textbf{Representation Aggregator.}
Messages transmitted from other vehicles need to be fused and interpreted by the ego vehicle. The Representation Aggregator (RA) for cooperative perception is implemented as a voxel max-pooling operation and a point transformer block. RA first spatially transforms the keypoints in other vehicles' coordinates to the ego vehicle's frame using their relative poses. This operation assumes accurate vehicle localization (\eg, using HD maps). It then aggregates the incoming messages that are spatially close via max-pooling all the points located inside the same voxel grid cell. Finally, it fuses the multi-view perception information with another Point Transformer block. The two operations above preserve the permutation invariance with respect to the ordering of other vehicles and can handle a variable number of sharing vehicles. For bandwidth control, \sysname receives messages from three randomly chosen V2V vehicles in the vicinity.
%The representation aggregator for cooperative perception is realized by a voxel max-pooling operation and another point-transformer layer. First, we assume perfect localization of vehicles, so we are able to precisely transform the key points in other vehicles' coordinates to the ego frame and put all the received key points along with ego key points into a ``bucket of points'' in the receiver's ego frame. We then merge the received messages that are spatially close to each other by max-pooling all the points in the same 3D space grid. Then we aggregate the multi-view perception information using another block of point transformer. Note that the two operations described above naturally preserve the permutation invariance of the ordering of other vehicles and is agnostic to the number of sharing vehicles. For bandwidth control, we randomly connect to 3 of the 6 nearest neighbors who are designed to have the sharing capability.

\vspace{1mm}
\noindent\textbf{Control Module.}
The control module is a fully-connected neural network designed to make control decisions based on the received messages. These control decisions include the throttle, brake, and steering, denoted as scalar $T,B,S$ respectively. 
% and the PID speed oracle has decision on brake $B_p$. 
These values output from the model are first clipped to their valid ranges (\eg, [0,1] for throttle). 
% When interacting with the environment, the agent has to keep moving towards a given target location within a speed limit. 
%To comply with the speed limit rules, we apply a PID controller and use the output (denoted as $B_p$) as a reference for brake. 
To comply with the speed limit rules, we apply a PID speed controller to prevent speeding. 
%The control decision $(a_\text{throttle}, a_\text{brake}, a_\text{steer})$ follows:
%\vspace{-3mm}
%\begin{equation*}
%    \begin{split}
%    a_\text{throttle} 
%    &=
%    \begin{cases}
%    0 
%    &\mbox{   if  } T \leq B \mbox{  or  } B_p > 0 \\
%    T  &\mbox{    otherwise}
%    \end{cases} \\
%    a_\text{brake} 
%    &= 
%    \begin{cases}
%    B & \mbox{   if  } T \leq B \\
%    B_p & \mbox{   if  }T>B \mbox{  %and  } B_p > 0\\
%    0 & \mbox{    otherwise}
%    \end{cases} \\
%    a_\text{steer} &= S        
%    \end{split}
%\end{equation*}    
%\vspace{-6mm}
\subsection{Policy Learning}
We train our model to imitate the expert policy with privileged information using DAgger~\citep{ross2011reduction}. To warm-start policy learning, we first train the model using behavior cloning.  

\vspace{1mm}
\noindent\textbf{Behavior Cloning.}
Behavior Cloning is designed to minimize the distribution gap between the training policy and the expert policy. The goal is to find an optimal policy $\hat{\pi}$ such that the loss w.r.t. the expert's policy $\pi_\text{expert}$, under its induced distribution of states $S$ is minimized, \ie,
\begin{equation}
    \hat{\pi} ={\arg\min}_{\pi \in \Pi} \mathbb{E}_{s \sim S}[\ell_\text{control}( \pi(s), \pi_\text{expert}(s))].
\end{equation}
The objective function $\ell_{\text{control}}$ is a linear combination of $\ell_1$-loss of throttle, brake, and steering between the policy's actions and the expert's actions:
\begin{equation}
    \ell_{\text{control}} 
    =\eta_1 \ell_{\text{throttle}} 
    +\eta_2 \ell_{\text{brake}} 
    +\eta_3 \ell_{\text{steer}} 
    \label{eq:control_loss}
\end{equation}
where $\eta_1, \eta_2, \eta_3$ are the coefficients of the loss for each action. All three coefficients are set to $1$ in our experiments. 

% \paragraph{DAgger.}
\vspace{1mm}
\noindent\textbf{DAgger.}
\label{dagger}
Limitations of behavior cloning for autonomous driving have been discussed in \citet{codevilla2019exploring}. DAgger~\citep{ross2011reduction} address the covariance shift issues via online training. The core idea is to let the student policy interact with the environment under the supervision of the expert and record the expert's actions on the same states visited by the student. The training dataset is iteratively aggregated, using a mixture of the student's and expert's actions. 
% the expert's action. The new trajectories are generated by a sampling policy, mixing the student's and expert's actions. 
The sampling policy $\pi_i$ for the $i$-th iteration follows:
\begin{equation}
        \pi_i = 
        \begin{cases}
            \pi_{\text{expert}}, & \mbox{w.p.   } \beta_i \\
            \pi_{\text{student},i}, & \mbox{w.p.    } 1-\beta_i
        \end{cases}
\end{equation}
where $\beta_i = \beta_0 \times \beta_{i-1}$ are exponentially decreasing from the initial $\beta_0$, representing the probability that the expert's action is executed at the $i$-th iteration.

\subsection{Implementation Details}
\label{sec:impl_detail}
% For the successful reproduction of our work, we specify implementation details in this section.

\vspace{1mm}
%When the ego vehicle receives messages from more than three neighboring vehicles, we randomly select messages from three of the vehicles. 
When more than three neighboring vehicles send messages, we randomly select messages from three of the vehicles. 
% According to our wireless profiling experiments (\S\ref{sec:autocast}), vehicles accept messages only from a stable radio range of 50 m.
% within 40 m with sharing capabilities to receive messages from them. %This neighbor selection process is not optimal but offers a good chance of messages containing information about the occluded areas. 
% Every raw LiDAR point cloud may contain up to 65536 coordinates.
%To down-sample the point cloud, we merge the points through max-pooling over a voxel grid of size $0.5m \times 0.5m \times 0.5m$. After ground plane removal, we randomly select 2048 points as the encoder input. 
%all the neighbors encode their processed point clouds locally by the 3-block Point Encoder and send the message of size $128 \times (128, 3)$ and warp the coordinates to the ego frame.
All the neighbors encode their processed point clouds locally by the 3-block Point Encoder and send the messages of size $128 \times (128, 3)$ and warp the coordinates to the ego frame.
%For three or more neighbors sending messages, the received message size is $384 \times (128, 3)$. We filter out the out-of-range messages and apply voxel max-pooling to these messages for size reduction. 
We aggregate the merged representations by another block of Point Transformer. After global max pooling, the features are concatenated with the ego speed feature before passing to the fully connected layer. 
%The multi-head attention layers in the Point Transformer uses an embedding size of 32, and the kNN considers 16 nearest neighbors.
%Our model has a 90ms latency on an NVIDIA GTX3090 GPU, where the point encoder takes 80ms.

Our model has a 90ms latency on an NVIDIA GTX3090 GPU, where the point encoder takes 80ms.
Our model training consists of two stages: behavior cloning and DAgger. We first train every scenario-specific model by behavior cloning, then the final policy of behavior cloning serves as an initial student policy for DAgger. We collect 4 new trajectories and append them to the Dagger dataset every 5 epochs using a sampling policy (see \S\ref{dagger}) with $\beta_0=0.8$ during the DAgger stage. For all data used for training, 25\% of them are collected under accident-prone scenarios (with an occluded collider vehicle inserted) and 75\% of them are normal driving trajectories.
%Our model training consists of two stages: behavior cloning and DAgger. We first train every scenario-specific model for 100 epochs by behavior cloning on 12 trajectories, with 3 of them collected under accident-prone scenarios (with an occluded collider vehicle inserted) and 9 of them being normal driving trajectories. The weights of neural networks are optimized with the Adam optimizer with an initial learning rate $10^{-4}$, weight decay $10^{-5}$, and batch size 32. The final policy of behavior cloning serves as an initial student policy for DAgger.  We collect 4 new trajectories every 5 epochs using a sampling policy (see \S\ref{dagger}) with $\beta_0=0.8$. The newly-collected data, which has a normal-to-dangerous ratio of 3:1, are aggregated to the DAgger training dataset. We run DAgger training for 105 epochs (\ie,~21 iterations). We evaluate the final policy after the last iteration. 
%The training time required varies among different models. The typical wall time needed for our model is about 60 hours with 24GB Nvidia GeForce 3090 GPU and 12 Intel(R) Core(TM) i9-10900KF CPU @ 3.70GHz CPUs. 
%
For more details, please refer to our supplementary materials and project website.
\section{\envname}
\label{sec:autocast}
We present \envname, a simulation framework which offers network-augmented autonomous driving simulation on top of CARLA~\cite{dosovitskiy2017carla}. This simulation framework allows custom designs of various traffic scenarios for training and evaluating cooperative driving models. The simulated vehicles can be configured with realistic wireless communications. It also provides a path planning-based oracle expert with access to privileged environment information.

% \subsection{CARLA}
% CARLA\cite{dosovitskiy2017carla} is a 3D driving simulator that allows users to simulate, train, and verify driving policies. The advantage of CARLA lies in its flexible sensor APIs and vivid urban driving scenes, which reduces the sim2real gap and avoid collecting expensive real-world data.  
% \subsection{AVR}

%Augmented Vehicular Reality(AVR) \cite{qiu2018avr} is a library for vehicle-to-vehicle communication built upon CARLA that allows simulated vehicles within a certain radius to form an team, communicate the perceived sensor data and perform perform early stage LiDAR fusion. A larger variety of data baggage can be transmitted by user design. Figure \ref{fig:lidar-fusion} shows the bird-eye-view fused-LiDAR data and single-LiDAR data sampled in the same run and at the same time stamp. The fused LiDAR data contains more information(red part) that is originally occluded. 

\begin{figure}
     \centering
    %  \begin{subfigure}{0.666\columnwidth}
    %      \centering
    %      \includegraphics[width=\textwidth]{Figures/scen6_bgtraffic.png}
    %      \caption{Overtake}
    %      \label{fig:overtake}
    %  \end{subfigure}
    %  \hfill
    %  \begin{subfigure}{0.666\columnwidth}
    %      \centering
    %      \includegraphics[width=\textwidth]{Figures/scen8_bgtraffic.png}
    %      \caption{Unprotected Left-turn}
    %      \label{fig:leftturn}
    %  \end{subfigure}
    % %  \caption{Example of benchmark scenarios}
    %  \label{fig:scenarios}
    %  \hfill
    %  \begin{subfigure}{0.666\columnwidth}
    %      \centering
    %      \includegraphics[width=\textwidth]{Figures/scen10_bgtraffic.png}
    %      \caption{Intersection}
    %      \label{fig:intersection}
    %  \end{subfigure}
    %  \caption{Example of benchmark scenarios}
    %\includegraphics[width=\columnwidth]{Figures/autocast_examples.pdf}
    \includegraphics[width=\columnwidth]{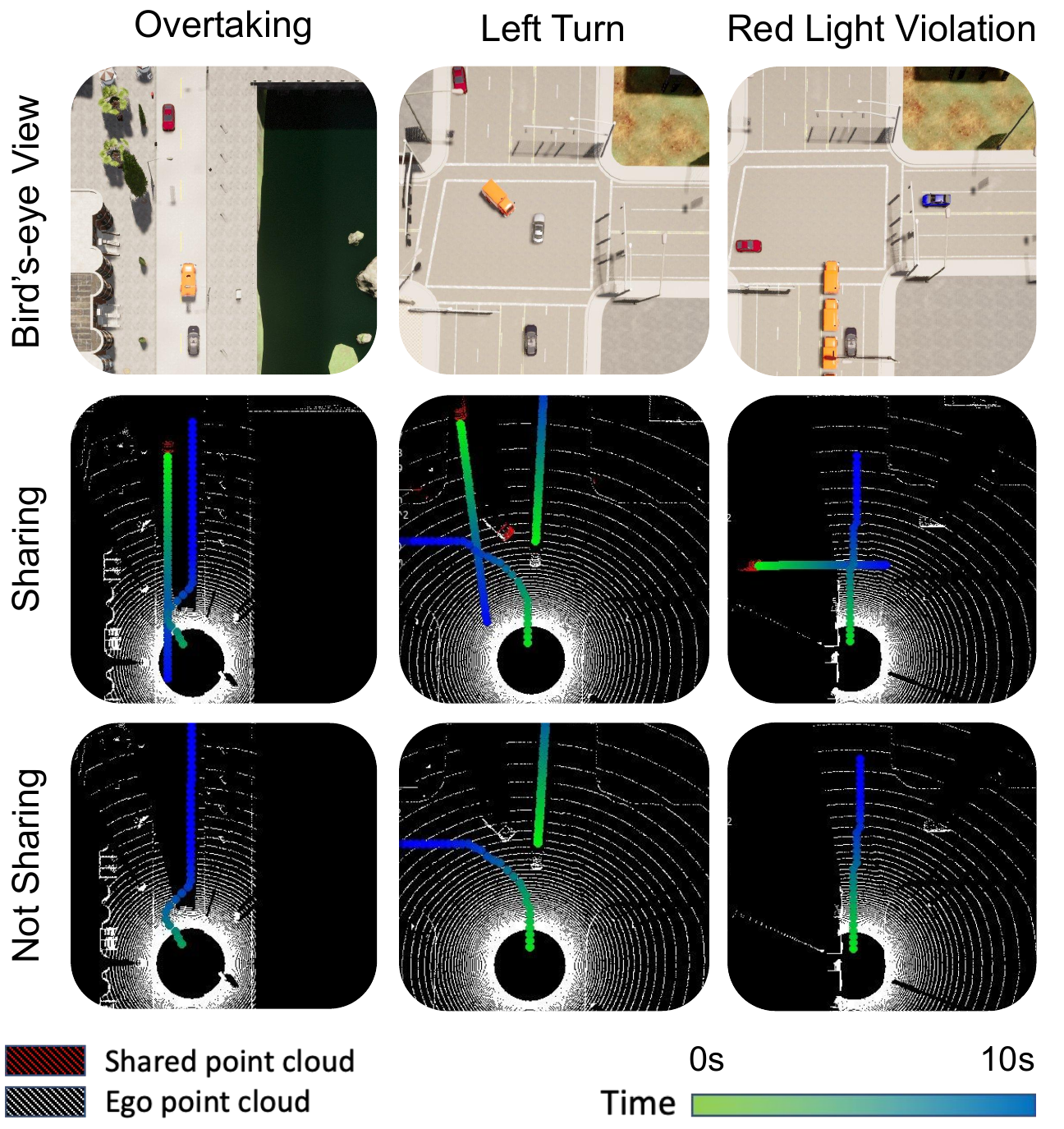}
    \vspace{-6mm}
    \caption{Benchmarking scenarios in \envname. The grey car is the ego vehicle  controlled by our model. The orange trucks are large vehicles that partially block views of the environment. The red car is not networked and likely to collide with the ego vehicle. All other vehicles are background traffic either with or without sharing capability. The green-blue dots mark the planned temporal trajectories for any moving vehicle, with green dots being waypoints closer in the future than the blue dots. If two planned trajectories intersect at a similar color (time), it indicates that a collision may happen. For every scenario, an RGB bird's-eye view (BEV), an ego-centric LiDAR BEV image, and a multi-vehicle fused LiDAR BEV image are presented (left to right). We use less background traffic here for illustration, and will study the effect of traffic density in \S\ref{sensitivity}.
    } 
    \label{fig:scenarios}
    \vspace{-5mm}
\end{figure}

%\changed{add legend, green->blue, one figure caption, 3 sub caption}

\subsection{Scenarios}
\label{sec:scennarios}
We designed three challenging traffic scenarios, shown in Figure~\ref{fig:scenarios}, in \envname as our evaluation benchmark. These scenarios are selected from the pre-crash typology of the US National Highway Traffic Safety Administration (NHTSA)~\cite{NHTSA-precrash}, where limited line-of-sight sensing affects driving decisions:

% . In these scenarios, cooperative perception can play an important role to impact . 
% there may be a vehicle in the occluded areas colliding with the ego vehicle. 
% Users can adjust the complexity by configuring different levels of background traffic for each run. 

%\begin{figure}
%    \centering
%    \includegraphics[width=.4\linewidth]{Figures/scen6_bgtraffic.png}
%    \includegraphics[width=.4\linewidth]{Figures/scen10_bgtraffic.png}
%    \caption[width=.9\linewidth]{Example Scenarios in the benchmark}
%    \label{fig:scenarios}
%\end{figure}

% \begin{itemize}
%     \item 
\vspace{1mm}
\noindent\textbf{* Overtaking.} A  truck blocks the way of a sedan in a two-way single lane road with a dashed yellow lane divider. The truck also impedes the sedan’s view of the opposite lane. The ego car has to overtake with a lane change maneuver.

    % \item 
    \vspace{1mm}\noindent
    \textbf{* Left Turn.} The ego car tries to turn left on a left-turn yield light but encounters another truck in the opposite left-turn lane, blocking its view of the opposite lanes and potential straight-driving vehicles.
    
    % \item 
    \vspace{1mm} 
    \noindent\textbf{* Red Light Violation.} The ego car is crossing the intersection when another vehicle is rushing the red light. LiDAR fails to sense the other vehicle because of the lined-up vehicles waiting for the left turn.
% \end{itemize}

%\subsubsection*{Other Scenarios}
%We also developed a parking scenario and left-turn yielding scenario. Please refer to our benchmark to be released for details. 

%a left-turning truck and a straight-going sedan is trying to cross the intersection under green light, when another sedan rushes a yellow light and/or violates a red light on the perpendicular way. The truck can see the violator and stops to avoid collision. But it is blocking the sedan’s view. An autonomous agent has to cross the intersection without collision.

\subsection{V2V Communication}
To simulate realistic wireless communication, we use real V2V wireless radios to profile wireless bandwidth capacity and packet loss rate due to channel diversity between mobile agents. Specifically, we use three iSmartways DSRC radios and three C-V2X radios~\cite{ismart}, mounted on top of moving vehicles, to measure the maximum capacity of continuous wireless transmission in practice. Table~\ref{tab:v2vcomm} shows the tested throughput and packet loss. It also shows the throughput of WiFi (802.11n, ac) for context. Note that the 802.11 series is not designed for mobile scenarios. Table~\ref{tab:v2vcomm} shows that V2V bandwidth is two orders of magnitude smaller than the indoor wireless capacity. The extremely limited bandwidth, in practice, poses significant challenges for designing the representations for V2V communication. We use the Winner II wireless channel model~\cite{meinila2009winner} in our simulator and use the measured 
%DSRC
C-V2X
radio capacity and packet loss rate in the channel model. We refer to prior work~\cite{autocast} for the design and implementations of the coordination, scheduling, and the network transport layer.

\begin{table}[!t]
\small
    \centering
    \resizebox{\columnwidth}{!}{
    \begin{tabular}{l|cc|cc}
    \toprule
		& DSRC & C-V2X & 802.11n & 802.11ac \\
		\midrule
        Throughput (Mbps) & 2.0 &  7.2 & $\sim$ 200 & $\sim$ 900\\
        Packet Loss (\%) &	$<5$ & $<5$ & $>90$ &	$>90$\\
        Mobility support & Yes &Yes & No & No\\
    \bottomrule
    \end{tabular}}
    \vspace{-2mm}
    \caption{Measured wireless throughput and packet loss rate using off-the-shelf wireless radios.}
    \vspace{-6mm}
    \label{tab:v2vcomm}
\end{table}
\subsection{Oracle Expert}
\label{sec:oracle}
The expert has access to the privileged information of the traffic scenarios. The information includes the point cloud from the LiDARs of all neighboring vehicles and the positions and speeds of these neighboring vehicles and other traffic participants. The expert transforms all of the point clouds from neighboring cars to its ego perspective (which is impractical due to the wireless bandwidth limit mentioned above). The transformed point cloud is fused for downstream obstacle detection and planning. The expert policy leverages all information above to analyze and avoid possible collisions. The path planning algorithm uses an A* trajectory planner~\citep{a_star} with pose and distance heuristics. The expert moves at a target speed of $20$km/h.

%Among the 6 collected trajectories, 3 of them do not have a rule-violating vehicle. These trajectories can bring ambiguity to the observation based on the single LiDAR agent, as the agents have no clue if there is a red-light-rusher until a very short time before collision. By creating such ambiguity, hopefully we will get some different policies from single LiDAR input and fused LiDAR input with such ambiguity.

%As for data pre-processing, we transform the 3D unstructured lidar data to voxel occupancy 3D grid with a width of $0.5m$. The ranges of each dimension are $[-70m, 70m], [-70m, 70m], [-2.5m, 2.5m]$ for dimension $X,Y,Z$ respectively.

%The supervised output is the next target location for local planner and brake control value, the general shape of desired output is $(x,y,z,\mbox{brake\_value})$. We filter out the abnormal frames where the output of teacher policy is missing.
\section{Experiments}
\vspace{-1mm}
We first discuss the evaluation method and the experiment setup and then give a brief overview of our baselines. Next we present the main quantitative evaluation results of our methods against baselines. Finally, we provide further analysis and visualization to understand the quality of our learned model. 

\subsection{Experimental Settings}
\label{sec:eval_method}
% \paragraph{Scenario Configuration.} 
%\vspace{1mm}
\noindent\textbf{Scenario Configuration.} 
We generate traces from the three scenarios we implemented in \envname (\S\ref{sec:scennarios}) for training and evaluation. These scenarios can be programmatically re-configured with key parameters, notably the number of vehicles, vehicle spawning locations, and vehicle cruising speeds. Random combinations of these parameters are sampled to procedurally generate traces with traffic situations of varying complexity --- in some cases the ego vehicle has to take emergency actions to avoid potential collisions, while in other cases,  cruising along the default route can reach the destination.
% \changed{double check the number of traces used}

\begin{table*}[h]
    \centering
    \resizebox{\textwidth}{!}{
    \begin{tabular}{lcccccccccc}
        \toprule
        \multirow{2}{*}{Model}
        &\multirow{2}{*}{Bandwidth}
        &\multicolumn{3}{c}{\textbf{Overtaking}} 
        &\multicolumn{3}{c}{\textbf{Left Turn}}
        &\multicolumn{3}{c}{\textbf{Red Light Violation}}\\
        \cmidrule{3-11}
        & (Mbps) 
        & SR$\uparrow$ & SCT$\uparrow$ & CR$\downarrow$ 
        & SR$\uparrow$ & SCT$\uparrow$ & CR$\downarrow$ 
        & SR$\uparrow$ & SCT$\uparrow$ & CR$\downarrow$ \\
        \midrule
        No V2V Sharing & -  
        & 45.3$\pm$0.6 & 43.6$\pm$0.7 & 35.8$\pm$3.6
        & 40.3$\pm$5.9 & 37.8$\pm$4.6 & 55.6$\pm$9.6
        & 47.3$\pm$18.7 & 46.1$\pm$18.4 & 51.4$\pm$17.4\\
        Early Fusion & 60.0 
        & 81.9$\pm$7.2 & 81.2$\pm$5.2 & 11.9$\pm$5.1
        & 72.8$\pm$8.6 & 68.8$\pm$8.9 & 26.3$\pm$8.1
        & 78.6$\pm$11.8 & 75.8$\pm$9.1 & \textbf{17.7$\pm$15.2}\\
        \midrule
        Voxel GNN & 5.60 
        & 70.0$\pm$4.8 & 67.8$\pm$4.2 & 16.1$\pm$3.6
        & 53.5$\pm$6.9 & 51.0$\pm$6.9 & 33.3$\pm$7.3
        & 64.2$\pm$25.3 & 62.0$\pm$24.8 & 35.0$\pm$25.9 \\
        \sysname (Ours) & \textbf{5.10} 
        & \textbf{90.5$\pm$1.2} & \textbf{88.4$\pm$1.1} & \textbf{4.5$\pm$3.1} 
        & \textbf{80.7$\pm$5.2} & \textbf{76.2$\pm$3.9} & \textbf{18.1$\pm$6.2}
        & \textbf{80.7$\pm$7.6} & \textbf{77.8$\pm$7.0} & \textbf{17.7$\pm$7.8}\\
        \bottomrule
    \end{tabular}
    }
    \vspace{-2mm}
    \caption{Quantitative results of different models over three repeated runs. SR: Success Rate, in percentage; SCT: Success weighted by Completion Time, in percentage; CR: Collision Rate, in percentage; In the Bandwidth column, we report the communication throughput required without data compression. The bandwidth is calculated by assuming 10 Hz LiDAR scanning frequency.}
    \label{tab:success_rate}
    \vspace{-4mm}
\end{table*}

% \paragraph{Dataset.} 
\vspace{1mm}
\noindent\textbf{Dataset.} 
Specifically, for each scenario, we use the expert agent (\S\ref{sec:oracle}) to generate an initial training set of 12 traces with randomized scenario configurations, followed by another randomly configured 84 traces for DAgger. 
In the evaluation, we systematically test each model on a spectrum of 27 randomly selected accident-prone environment configurations over three repeated runs, each using different random seeds for background traffic. For fair comparison, we use a fixed set of 27 test configurations to evaluate all models.

% Notice that, since the normal driving situation is most likely to be safe, we mix the normal driving trajectories with those with emergent situations during training, keeping the ratio at 3:1.

% We train separate policies for each scenario under randomized background traffics and randomly sampled environment configurations. Then we test the policies for a spectrum of 27 dangerous environment configurations, with a fixed random seed for background traffic, so that the difficulty presented to the different agents are the same. We report the success rate and mean success weighted by completion time (defined later in this section) with a paired comparison to expert trajectories.

% \paragraph{Metrics.} 
\vspace{1mm}
\noindent\textbf{Metrics.} 
We report three metrics, \textit{Success Rate}, \textit{Collision Rate}, and \textit{Success weighted by Completion Time}:

\vspace{1mm}
\noindent\textit{Success Rate (SR).} 
% Multiple criteria are considered during evaluations. 
A \textit{successful completion} of the scenario is defined as the ego agent reaching a designated target location in a permissible time without collision or prolonged stagnation. The success rate is defined as the percentage of \textit{successful completions} among all evaluated traces.
% \commentp{Do we ever specify the permissible time or allowed stagnation?}
%\subsubsection*{Route Completion} It is designed to incorporate the situation where a scenario is too hard to complete, then we look at the intermediate result, the route completion rate. A mean route completion rate is reported on the entire test settings. 

\vspace{1mm}
\noindent\textit{Collision Rate (CR).} 
Collision is the most common failure mode. \textit{Collision rate} is defined as the percentage of evaluation traces where the ego vehicle collides with any entity, such as vehicles, buildings, etc.

\vspace{1mm}
\noindent\textit{Success weighted by Completion Time (SCT).}
SR reflects overall task success or failure. It does not differentiate the amount of time a driving agent needs to complete the traces. We introduce a third metric to weigh the success rate by the completion time ratio between the expert and the agent:
% \changed{More cautious non-V2V
% driving could crash less. However, blindly acting cautiously
% will also hinder the efficiency in routing driving, especially
% as accident-prone scenarios are rare.}
% We propose the success weighted by completion time to measure the completion time of policies.
% \jiaxun{(Choose between these two sentences) Given the same source location and target location, a bad policy may be too conservative at making decisions, resulting in prolonged stagnation or too aggressive, resulting in accidents.} The SCT metric is defined as follows:

%\begin{equation}
%    SCT =\mathbb{I}\{\text{agent success}\}\frac{T_\text{expert}}{T_\text{agent}}
%\end{equation}
\begin{equation}
    SCT =\mathbb{I}\{\text{agent success}\}{T_\text{expert}}/{T_\text{agent}}
\end{equation}
where $\mathbb{I}$ is an indicator function, and $T_\text{expert}$ and $T_\text{agent}$ are the expert's and the agent's completion time, respectively. As the expert agent should require no longer completion time than the agent, the ratio resides in the range of $[0, 1]$.

\subsection{Baselines}
We compare \sysname with non-V2V and V2V driving baselines. For fair comparison, we adopt the same neighbor selection process (\S\ref{sec:impl_detail}) among V2V approaches:

% \begin{itemize}
%     \item 
    \vspace{1mm}
    \noindent
    \textbf{* No V2V Sharing.} The non-sharing baseline makes decisions solely based on the onboard LiDAR data and ego speed. 
    % No communication happens here. 
    %The model architecture is similar to our final model. Instead, we replace the representation aggregator with a regular point transformer block.
    The model shares the same data processing scheme for an individual vehicle and point encoder architecture as our final model.
    % \item 
    
    \vspace{1mm}
    \noindent
    \textbf{* Early Fusion.}
    The Early Fusion model assumes an unrealistic communication bandwidth, with which it can transmit and fuse the entire raw point cloud data from all neighboring vehicles. While this method is intractable in practice, it serves as a baseline to examine our point-based architecture's effectiveness in representation learning. 
    %Like the previous baseline, Early Fusion also uses a 3-block point transformer instead of the representation aggregator.
    To fit this model in GPU memory, we limit the size of the fused input points to 4,096. Like the previous baseline, Early Fusion also uses a 3-block Point Transformer encoder.
    
    % \item
    \vspace{1mm}
    \noindent
    \textbf{* Voxel GNN.} We adapt V2VNet~\cite{wang2020v2vnet}, which is designed for 3D detection and motion forecasting, to learn end-to-end driving. Every vehicle processes its local point cloud onboard and shares a voxel representation with the ego vehicle for control. It uses a graph neural network (GNN) in the ego frame as the aggregator. The control actions are predicted from the GNN-fused representations.

For fair comparison, all baselines and proposed approaches are independently trained over three repeated runs with the same training parameters (\S \ref{sec:impl_detail}). We report the average performance over the three runs on the same scenario configurations (\S \ref{sec:eval_method}).

% \subsection{Results}
\subsection{Quantitative Results}
This section presents the empirical evaluations of all the models in the three benchmarking scenarios.
% , a sensitivity analysis on the background traffic level and the number of neighbors.

\vspace{1mm}
\noindent\textbf{Scenario Completion.}
% \paragraph{Scenario Completion.}
Table~\ref{tab:success_rate} shows the performance comparisons in each of the three traffic scenarios. In all three scenarios, the No V2V Sharing model has performed poorly, with less than 50\% success rate for each scenario and high collision rates. All three cooperative driving models, including Early Fusion, Voxel GNN, and \sysname, have achieved substantially higher SR and SCT scores and lower collision rates than the 
No V2V Sharing baseline. It indicates that the V2V communication provides critical information about the traffic situation over the ego vehicle's line-of-sight sensing to make informed driving decisions. The Early Fusion method improves over the non-V2V baseline over 30\% in average success rate. However, the Early Fusion baseline requires transmitting raw point clouds across vehicles, leading to an unrealistic bandwidth requirement of $60$Mbps (before data compression).

In contrast, pre-processing raw sensory data into representations has dramatically reduced the bandwidth requirements while improving driving performances. Both Voxel GNN and \sysname perform sensory fusion on the representation level. 
%We observe that \sysname outperforms the other baselines on Overtaking and Left Turn, while Voxel GNN reports the strongest performance on Red Light Violation. We hypothesize that the ego vehicle can benefit more from cooperative perception in the Left Turn and Overtaking scenarios. In these scenarios, the ego vehicle must yield the right-of-way to opposite going cars during an unprotected left turn and lane changing. It has to detect the neighboring vehicles and exhibit highly reactive behaviors to other cars. \sysname outperforms 7\% over Voxel GNN and more than 40\% over the non-sharing baseline on average success rate.
In comparison to the other cooperative driving models, \sysname outperforms both Early Fusion and Voxel GNN baselines for all three scenarios. We hypothesize that the point-based representation learning of \sysname makes it robust to localization errors compared with fusing raw points in Early Fusion. Furthermore, the explicit representation of point 3D locations and the point sampling module of \sysname retain a high spatial resolution of its intermediate representations in contrast to the voxel-based feature maps used by Voxel GNN.
% Our experimental results show the necessity of vehicle-to-vehicle communication. We notice an inconsistency of best performer across different scenarios: while our method leads in 2 of the investigated scenarios, and Voxel-GNN achieves the best result in 1 of the scenarios. More interestingly, our representation fusion is constantly better than the early-fusion method in all scenarios with the same point processing backbone.  

\vspace{1mm}
\noindent\textbf{Bandwidth Requirement.}
As shown in Table~\ref{tab:success_rate}, sharing raw point cloud at the LiDAR scanning rate of 10fps would require a wireless bandwidth of 60Mbps, far beyond the achievable bandwidths in the current (DSRC) and future (C-V2X or LTE-direct) V2V communication technology (expected less than 10Mbps, see Table~\ref{tab:v2vcomm}). V2VNet~\cite{wang2020v2vnet} claims a bandwidth requirement of 25 Mbps with point cloud compression, which is also beyond what current V2V radios can support. In our design, both Voxel GNN and \sysname requires less than 6Mbps bandwidth, a 4$\times$ reduction of the communication data sizes of V2VNet without compression. 

When developing the V2V models, we carefully explored the design space of the sharable representation size and its bandwidth requirement for both Voxel GNN and \sysname. For example, if \sysname were to share a 32$\times$32 representation, it only needs 0.9 Mbps. However, the coarse information is insufficient for the model to attain a good performance. We find that a 128$\times$128 point representation meets the bandwidth requirements (Table~\ref{tab:v2vcomm}) without substantial performance degradation.

\vspace{1mm}
\noindent
\textbf{Sensitivity to Traffic Density.}
\label{sensitivity}
\begin{figure}
\centering
\includegraphics[width=0.99\columnwidth]{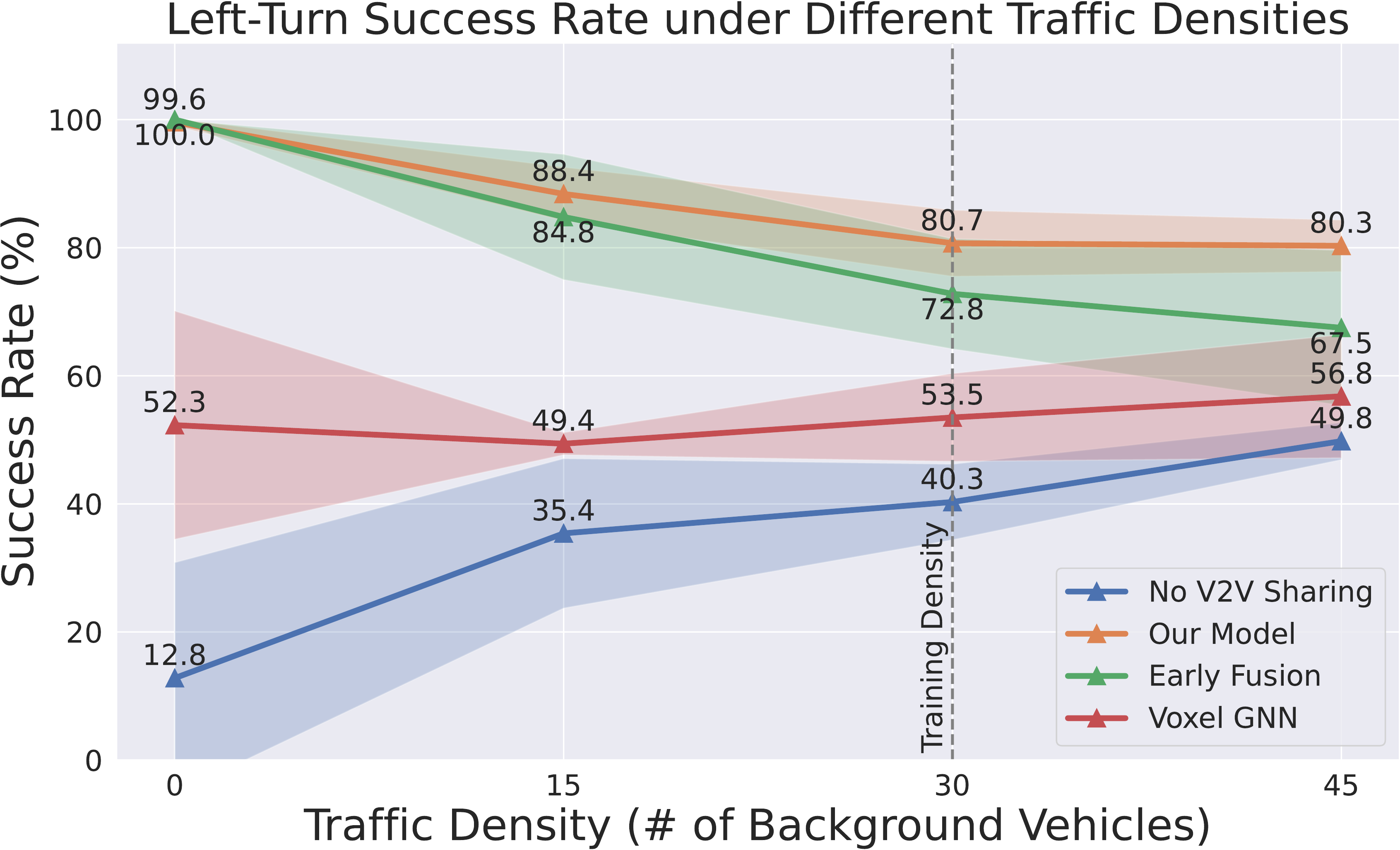}
\vspace{-2mm}
\caption{Sensitivity analysis on the varying levels of traffic densities in the Left Turn scenario.}
\vspace{-3mm}
\label{fig:sensitivity}
\end{figure} 
We further test \sysname under varied traffic densities in the most challenging Left Turn scenario. Figure \ref{fig:sensitivity} shows that our method generalizes to variable traffic densities, consistently outperforming the No V2V Sharing baseline. 
Qualitatively, we observe that No V2V Sharing drives slower in denser traffic,  reacting better to emergency situations. In contrast, V2V methods do not improve much in denser traffic, as they tend to be impacted by the increased stochasticity of incoming messages from changing neighbors. Nonetheless, \sysname outperforms the baselines in all traffic densities with over $30\%$ higher success rates than No V2V Sharing.

%Interestingly, the No V2V Sharing baseline has a higher success rate with increased traffic. It may be because the driving policy becomes more cautious and lowers speed in the presence of more traffic participants, thus less prone to collisions.  

%The collision behavior can be affected by the background traffic density, which is why the non-sharing method has an increasing success rate, while with less background traffic the useful messages are not dilated which is why our model has higher success rate when only the occluding vehicle is sharing (background traffic=0).

\begin{figure}
\centering
\includegraphics[width=0.5\textwidth]{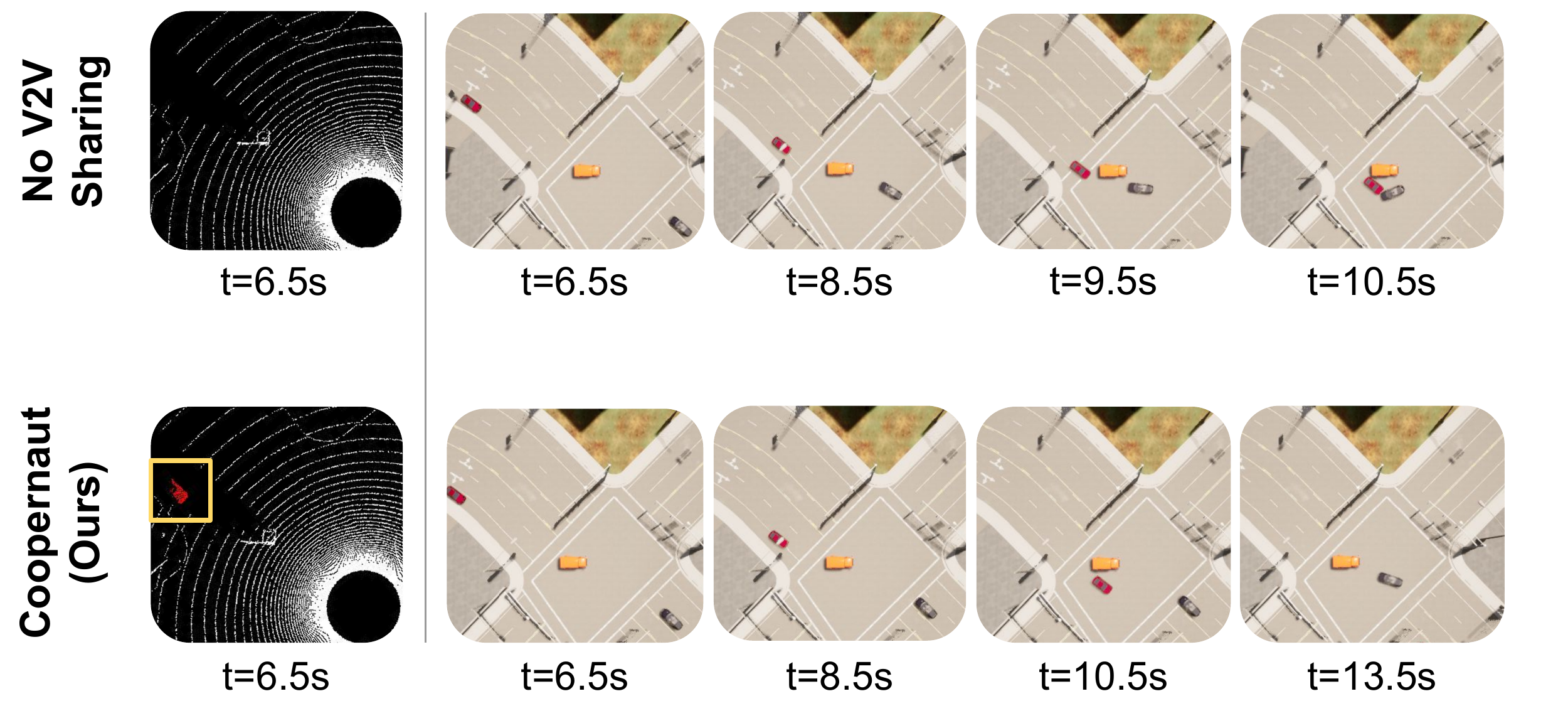}
\vspace{-5mm}
\caption{Comparison of trajectories in the Left Turn scenario. The grey car in the pictures is the controllable ego vehicle. The red car is going straight in the opposite direction, occluded behind the orange truck. Our model avoids the collision as it is able to see the red light violating vehicle from cooperative perception (highlighted in the yellow box).}
\label{fig:trajectory}
\vspace{-3mm}
\end{figure}

\vspace{1mm}
\noindent\textbf{Qualitative Visualizations.}
% \paragraph{Qualitative Visualizations.}
Figure \ref{fig:trajectory} shows an example evaluation trajectory from Left Turn. The left-turning ego vehicle (grey) can proactively avoid collision by yielding to the opposite-going cars with \sysname.
A common failure pattern of the non-sharing model is that it drives ahead to its target location regardless of any traffic violators or potential colliders due to the limited line-of-sight of its ego perception. %Since the oracle has global information, it will brake when there is a potential collision, but accelerate ahead when there is not. However, the LiDAR observation of non-sharing mode in these two situations is similar: the collider is invisible. Therefore, the ambiguity in the control decision confused the learned policy, which results in failure. In contrast, 
The transmitted messages through V2V channels help our model resolve the ambiguity with cross-vehicle perception, leading to safer driving decisions in this accident-prone situation.

% and effectively extract the critical yet invisible information, compile them into compact representations for sharing, and efficiently fuse these representations to reason about the dangerous situation.

% Though there are traces in collected training set, which illustrates how the oracle expert react to the emergent situation,  existence of the ambiguity in the LiDAR observation and distracting normal driving trajectories cause the failures in learning reactions to the dangerous scenarios. But the our model with communicated perception can capture these rare cases and help distinguish them from normal driving data. 

\subsection{Limitations and Future Work}
While our cooperative perception model conforms to realistic wireless bandwidth, we do not take into account practical networking issues, including transmission latency, networking protocols, and repetitive or lost packets. Nonetheless, \sysname is robust to packet loss to a certain extent ($5\%$ as configured in \envname). Its random neighbor selection also adds another layer to endure packet loss from individual transmitters. 

Furthermore, highly accurate vehicle localization is assumed, which is used by \sysname to transform the point-based representations from neighboring vehicles to the ego vehicle, even though \envname simulates slight errors in the pose and height estimation of a vehicle. In reality, without a high definition map (HDMap), localization error can yield up to meter-level displacement. Using HDMap can significantly improve location and pose estimation, which is commonly adopted in both industry and academia~\cite{yang2020hdnet, li2021hdmapnet}. 

% Our model has only been tested in the specific domains, and learning only from ego control can cause the issue of over-fitting to the environment. It is also expected that the encoder needs significantly large number of samples to learn a generalizable features. Second, as perfect localization is assumed, it may be the case that our model will suffer from the localization error. One way to test it is to add random noise to the pose of every vehicles to bridge the sim-2-real gap. 

For fair comparison, we use the same down-sampling scheme for all point-based baselines and our approach, which proves to be effective in our scenarios with moving vehicles and large obstacles. For smaller objects like pedestrians, adaptive sampling schemes based on semantic information is a promising direction for future work. We would also like to extend the model architecture of \sysname to better incorporate temporal information for improving driving performance.

\section{Conclusion and Future Work}
\vspace{-1mm}

This work investigates vision-based driving using cooperative perception for networked vehicles in a newly designed simulation benchmark \envname. We introduce \sysname, an end-to-end driving policy that encodes, aggregates, and analyzes 3D LiDAR data from networked vehicles. The point encoder and representation aggregator of \sysname retain detailed spatial information and are robust to a varying number of communicating vehicles. Our empirical results show that our method improves the robustness of autonomous driving policies in risk-sensitive traffic scenarios.

This work has ample room for future extension. Our method relies on a hand-engineered oracle for imitation learning. It leaves open questions to investigate adaptive strategies of \textit{when} to communicate, \textit{what} to encode in messages, and \textit{how} to drive cooperatively, ideally without the need of an algorithmic oracle.

% We currently train separate models for different scenarios, and the model only receives . It is interesting to investigate the reinforcement learning aspect of the driving policies with cooperative perception. We are also interested in moving forward to a multi-agent decision-making process and improving the scalability of \envname benchmark.

\vspace{-2mm}
{\small
\paragraph{Acknowledgements}
 This work has taken place in the Robot Perception and Learning Lab (RPL) and Learning Agents Research Group (LARG) at UT Austin. RPL research is supported in part by NSF (CNS-1955523, FRR-2145283), the MLL Research Award from the
Machine Learning Laboratory at UT-Austin, and the Amazon Research Awards. LARG research is supported in part by NSF (CPS-1739964, IIS-1724157, FAIN-2019844), ONR (N00014-18-2243), ARO (W911NF-19-2-0333), DARPA, Lockheed Martin, GM, Bosch, and UT
Austin's Good Systems grand challenge.  Peter Stone serves as the
Executive Director of Sony AI America and receives financial
compensation for this work.  The terms of this arrangement have been
reviewed and approved by the University of Texas at Austin in accordance
with its policy on objectivity in research.}

% The acknowledgments are automatically included only in the final and preprint versions of the paper.
% \acknowledgments{If a paper is accepted, the final camera-ready version will (and probably should) include acknowledgments. All acknowledgments go at the end of the paper, including thanks to reviewers who gave useful comments, to colleagues who contributed to the ideas, and to funding agencies and corporate sponsors that provided financial support.}

%===============================================================================

% \bibliography{reference}  % .bib
% \bibliographystyle{IEEEtranS}
% \newpage
\bibliography{reference}

\end{document}